\documentclass[10pt,twocolumn,letterpaper]{article}

\usepackage{cvpr}              

\usepackage{graphicx}
\usepackage{kotex}

\usepackage{makecell}
\usepackage{multirow}
\usepackage{adjustbox}

\newcounter{JYNumberOfComments}
\stepcounter{JYNumberOfComments}

\newcounter{YSNumberOfComments}
\stepcounter{YSNumberOfComments}

\definecolor{cvprblue}{rgb}{0.21,0.49,0.74}
\usepackage[pagebackref,breaklinks,colorlinks,allcolors=cvprblue]{hyperref}

\def\paperID{18} 
\def\confName{CVPR}
\def\confYear{2025}

\title{KOFFVQA: An Objectively Evaluated Free-form VQA Benchmark for Large Vision-Language Models in the Korean Language}

\author{Yoonshik Kim $^{*}$ \quad Jaeyoon Jung \thanks{
    Equal Contribution.
}\\
MAUM AI Inc. / Republic of Korea \\
{\tt\small{\{yoonshik1205, jyjung\}@maum.ai}}
}

\begin{document}
\maketitle
\begin{abstract}
The recent emergence of Large Vision-Language Models(VLMs) has resulted in a variety of different benchmarks for evaluating such models. Despite this, we observe that most existing evaluation methods suffer from the fact that they either require the model to choose from pre-determined responses, sacrificing open-endedness, or evaluate responses using a judge model, resulting in subjective and unreliable evaluation. In addition, we observe a lack of benchmarks for VLMs in the Korean language, which are necessary as a separate metric from more common English language benchmarks, as the performance of generative language models can differ significantly based on the language being used. Therefore, we present KOFFVQA, a general-purpose free-form visual question answering benchmark in the Korean language for the evaluation of VLMs. Our benchmark consists of 275 carefully crafted questions each paired with an image and grading criteria covering 10 different aspects of VLM performance. The grading criteria eliminate the problem of unreliability by allowing the judge model to grade each response based on a pre-determined set of rules. By defining the evaluation criteria in an objective manner, even a small open-source model can be used to evaluate models on our benchmark reliably. In addition to evaluating a large number of existing VLMs on our benchmark, we also experimentally verify that our method of using pre-existing grading criteria for evaluation is much more reliable than existing methods. Our evaluation code is available at \href{https://github.com/maum-ai/KOFFVQA}{\texttt{https://github.com/maum-ai/KOFFVQA}}.
\vspace{-1em}
\end{abstract}    
\section{Introduction}
\label{sec:intro}

In recent years, there has been much research on Large Vision-Language Models(VLMs), giving rise to models with unprecedented capabilities in various applications~\cite{liang2024survey, yin2023survey}. However, evaluating the performance of such generative models remains a difficult task~\cite{huang2024survey}.
Although there are many benchmarks for evaluating the performance of VLMs, they are limited by the fact that the performance of a generative language model is hard to measure quantitatively~\cite{chang2024survey}. Most benchmarks either restrict the model to respond with one of several pre-defined answers~\cite{fu2023mme, liu2025mmbench, chen2024we, yue2024mmmu, guan2024hallusionbench} or evaluate the outputs of the model using potentially subjective metrics such as LLM judges~\cite{yu2023mm, liu2023visual}.

An important property of generative language models, including VLMs, is that they have the ability for open-ended generation of long-form responses~\cite{yin2023survey}. When evaluating the general-purpose performance of such models, not being able to harness this ability can be limiting. However, evaluating long-form responses is very challenging~\cite{huang2024survey}. Many benchmarks use an LLM or VLM judge to evaluate such responses, but this can be subjective and unreliable~\cite{thakur2024judging, chang2024survey, wang2023evaluation}. This approach can also be biased due to the tendency of judge models to give comparatively higher scores to the responses of similar models~\cite{panickssery2025llm, zheng2023judging}.

In addition, grading a VLM benchmark using a judge model requires the judgment process to incorporate information from both images and language queries, unlike LLM benchmarks. This can be achieved through methods such as 1) using a VLM-as-a-judge approach and providing the images directly to the judge~\cite{chen2024mllm, lee2024prometheus}, 2) prompting the judge to evaluate the response by comparing it to a second baseline or ground-truth response~\cite{yu2023mm, liu2023visual}, or in the case of specialized benchmarks, 3) using some ground-truth information extracted from the image in the judgment process~\cite{Li-hallucination-2023}.
However, each of these methods has limitations. In the first approach, when VLMs are used to judge responses, such models can be prone to hallucinations based on visual input~\cite{wang2023evaluation} or can be misaligned with human intention~\cite{chen2024mllm}. In the second approach, comparison with a baseline or ground-truth response can be limiting since there may be more than one way to correctly answer a question. In the third approach, it is difficult to generalize such methods outside the domain of specialized benchmarks.

To address these challenges, we propose a novel approach to evaluating such open-ended responses in VLM benchmarks: providing a list of objective pre-defined grading criteria paired with partial scores to an LLM judge in order to reliably grade long-form responses to queries. Additionally, we propose that using a VLM as a judge and providing images to aid in judgment can negatively affect the consistency of evaluation when using our approach due to the tendency of VLMs to hallucinate based on visual input~\cite{wang2023evaluation} and that our approach does not suffer from this as it uses an LLM judge.

Furthermore, despite the fact that the performance of generative language models can differ greatly based on the language being used~\cite{chang2024survey}, there are very few benchmarks for VLMs in languages other than English or Chinese. In order to evaluate the performance of VLMs in other languages, such as Korean, there is a need for general-purpose benchmarks that can evaluate various aspects of the performance of a VLM in those languages.

Based on this, we introduce \textbf{KOFFVQA}, 
a carefully hand-crafted general-purpose visual question answering(VQA) benchmark for the Korean language capable of objectively evaluating free-form responses from models. Each question consists of an image, the corresponding question, and a list of pre-defined grading criteria for responses. For each given image-question pair, the response of a VLM is evaluated by an LLM judge that is simply instructed to score the response based on the given criteria. In addition, the tasks in our benchmark attempt to cover as many aspects of VLM performance relevant to real-world applications as possible. We propose that this benchmark does not suffer from the issues that affect the reliability of evaluating long-form responses, and that it fills the much-needed gap in general-purpose VLM benchmarks for the Korean language.
We evaluate 47 total open-source and proprietary VLMs on our benchmark, and also experimentally compare our methodology with existing approaches to VLM benchmarks. Our contributions are as follows:

\begin{itemize}
\item We create and release KOFFVQA, a general-purpose Korean-language VLM benchmark that uses question-specific grading criteria to objectively evaluate free-form responses. Our benchmark consists of 275 human-written questions each paired with an image and their corresponding grading criteria across 10 subcategories designed to evaluate various aspects of a VLM's performance.
\item We evaluate 47 open- and closed-source VLMs on our benchmark. We find that a larger model does not necessarily mean higher performance unlike in English-language benchmarks, and that models that excel in some subcategories do not necessarily excel in others.
\item We compare our approach of benchmark evaluation with the existing method of comparing each response with a baseline. We find that our approach significantly increases the consistency of evaluation compared to the existing method. Furthermore, the fact that our approach can utilize LLM-as-a-judge for evaluation mitigates the issues that may arise when using VLM-as-a-judge due to the tendency of VLMs to hallucinate based on visual input. We find that using the LLM-as-a-judge method is more reliable and accurate when using our approach compared to using the VLM-as-a-judge method with image input.
\end{itemize}
\section{Related Work}
\label{sec:relatedwork}

\subsection{VLM Evaluation Benchmarks}

There has been much work on benchmarks for evaluating large generative vision-language models~\cite{yin2023survey, fu2024mme, chen2024we, liu2025mmbench, yue2024mmmu, guan2024hallusionbench, liu2023visual, chen2024mllm, yu2023mm, singh2019towards, lu2023mathvista}. While there is a large variety of benchmarks that evaluate various aspects of performance, the open-ended nature of text generation makes it difficult for such generated responses to be evaluated using traditional metrics~\cite{chang2024survey}. Due to this, existing VLM benchmarks largely fall into the following two categories:

\noindent \textbf{Multiple Choice Question Answering.} These benchmarks provide a fixed set of possible answers to each question(including simply “yes” or “no”), and prompt the model to choose the correct answer. This is usually achieved by either prompting the VLM to generate a response in a format that can be easily parsed~\cite{chen2024we} or using a separate LLM to extract the answer choice from the VLM’s response~\cite{liu2025mmbench}. Such benchmarks include MME~\cite{fu2023mme}, MMBench~\cite{liu2025mmbench}, MMStar~\cite{chen2024we}, MMMU~\cite{yue2024mmmu}, and HallusionBench~\cite{guan2024hallusionbench}. This approach allows for objective evaluation, but is limited in that it cannot evaluate long free-form responses, which are characteristic of generative models.

\noindent \textbf{Free-form Question Answering with Subjective Evaluation.} This approach involves prompting the VLM to generate a free-form response to a question, then prompting a separate judge model(either an LLM or a VLM) to assign a score to each generated response, such as by comparing the generated response to a reference response~\cite{liu2023visual} or by grading the quality of the response using subjective criteria~\cite{chen2024mllm}. Such benchmarks include MM-Vet~\cite{yu2023mm} and LLaVA-Bench~\cite{liu2023visual}. This approach is useful for evaluating the quality of long free-form responses but suffers from the subjectivity of evaluation criteria, which can often lead to inconsistent or biased evaluation~\cite{zheng2023judging, shen2023large}.

There are some exceptions, most notably benchmarks that require the model to output an exact single word or phrase such as TextVQA~\cite{singh2019towards} and MathVista~\cite{lu2023mathvista}.

\subsection{Fine-grained Evaluation}

Using LLM- or VLM-as-a-judge methods for question answering with subjective evaluation has the aforementioned issues of inconsistent or biased evaluation. As such, there has been some work on mitigating this using fine-grained evaluation criteria.
Prometheus~\cite{kim2023prometheus} and Prometheus-vision~\cite{lee2024prometheus} are judge models that can evaluate the responses of a model on benchmarks on a scale of 1 to 5, given a single user-specified grading criterion. Evallm~\cite{kim2024evallm} provides a framework for evaluating and providing feedback on language model outputs based on user-given criteria. BiGGen Bench\cite{kim2024biggen} is a benchmark for LLMs that utilizes a single instance-specific criterion per instruction to evaluate the response of a model on a scale of 1 to 5. FLASK-Hard~\cite{ye2024flask} is a fine-grained evaluation protocol for LLMs that utilize instance-specific criteria for 12 skill sets that are used to evaluate the responses of LLMs.

However, these methods have limitations. First, adherence to the five-point scale makes the approach inflexible. For example, it is difficult to create grading criteria for simple questions that can only be correct or incorrect. Furthermore, the majority of the instance-specific criteria used in the above studies are subjective metrics, and are still open to judge alignment issues.
We believe our approach suffers from neither of these limitations, as it uses a flexible system for criteria and only uses criteria that can be evaluated objectively by the judge model. Additionally, this allows our approach to utilize even small open-source models for reliable evaluation, which makes the evaluation process much more accessible.

\subsection{Korean VLM Benchmarks}

The performance of generative language models can vary across different languages~\cite{chang2024survey}. Despite this, there has not been much work regarding VLM benchmarks in the Korean language. The VARCO-VISION~\cite{ju2024varco} paper introduces four open multiple-choice benchmarks(K-MMBench, K-SEED, K-MMStar, K-DTCBench), three of which are translated subsets of existing English language benchmarks. VLR-Bench~\cite{lim-etal-2025-vlr} is a VLM benchmark in English, Chinese, and Korean for evaluating retrieval-augmented generation ability. \cite{lee2024harnessing} introduces a Korean-English bilingual multiple-choice benchmark for specifically evaluating a VLM's ability to distinguish between animate and inanimate objects.
Moreover, VQA datasets in the Korean language that are not necessarily intended for the evaluation of VLM performance can be used as benchmarks, as in \cite{shin-etal-2024-x}. These include KVQA~\cite{kim2019korean}, which is a Korean language VQA dataset featuring situations commonly encountered by blind people, and BOK-VQA~\cite{kim2024bok}, a Korean-English bilingual VQA dataset for knowledge-based VQA.

We observe that most of these datasets are not necessarily general-purpose benchmarks designed to evaluate various aspects of a VLM's performance. To the best of our knowledge, the only general-purpose Korean-language VLM benchmarks publicly available are K-MMBench, K-SEED, and K-MMStar, all of which are translated subsets of English-language benchmarks and only utilize multiple-choice questions.
\begin{figure}[t]
\centering
\includegraphics[width=0.88\linewidth]{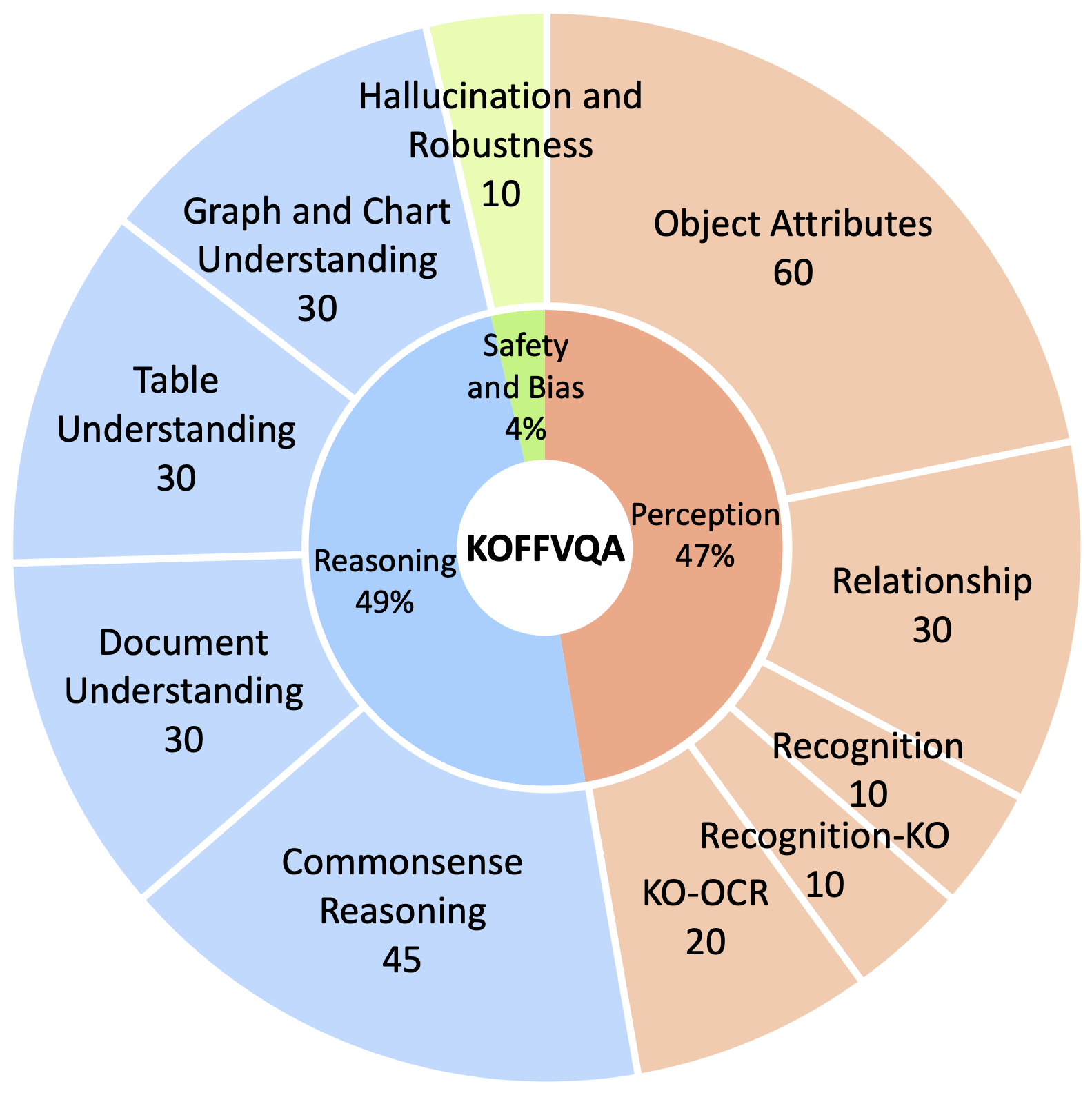}
\caption{Distribution of question categories and subcategories in the KOFFVQA benchmark.}
\label{tab:amount}
\end{figure}
\section{The KOFFVQA Benchmark}
\label{sec:benchmark}

\begin{figure*}[t]
\includegraphics[width=0.99\linewidth]{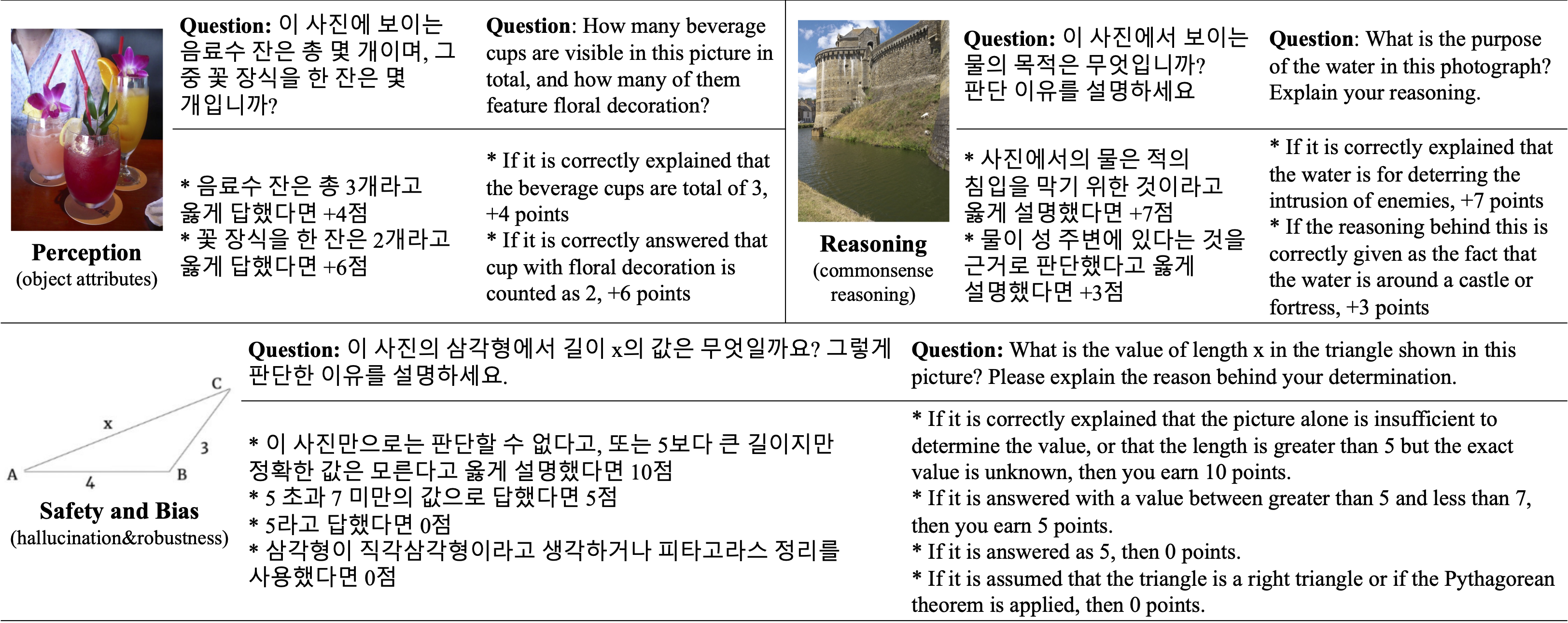}
\caption{Three examples from each main category of our benchmark. The left column is the original text in Korean, and the right column provides the English translation. Grading criteria paired with partial points are given to the judge model to evaluate the VLM's response.}
\label{tab:questionexample}
\end{figure*}

Existing VLM benchmarks that use the LLM- or VLM-as-a-judge approach may suffer from the problem of unreliable judgment, and even when utilizing fine-grained criteria, suffer from limitations associated with such approaches.
Therefore, we present KOFFVQA, a general-purpose Korean-language VLM benchmark with fine-grained objective evaluation criteria designed to evaluate a wide range of aspects of VLM performance relevant to real-world applications.

\subsection{Tasks}
In this benchmark, we define 3 main categories and 10 subcategories of questions based on the categorization schemes used by existing general-purpose datasets~\cite{chen2024we, liu2025mmbench, huang2024survey}. The categorization that we use is as follows:

\begin{enumerate}
\item \textbf{Perception}
    
    \textbf{Object Attributes.} These questions ask about the attributes of objects within an image, such as color, shape, count, or existence. This evaluates the most basic skill necessary to perceive and reason with visual input.
    
    \textbf{Relationship.} These questions ask about aspects of a complex scene involving multiple elements, such as attribute comparisons or spatial relationships. This evaluates the most basic skill necessary to reason about images that involve multiple visual elements.
    
    \textbf{Recognition.} These questions ask the model to recognize well-known visual subjects, such as the names of places, artworks, people, or symbols. This evaluates the level of visual world knowledge that a model possesses.
    
    \textbf{Recognition-KO.} We create a separate subcategory of recognition specific to Korean culture, where the questions ask about visual subjects that are specifically well-known among speakers of the Korean language. This evaluates the specialized skill necessary to understand scenes that may commonly be presented when being used with the Korean language.
    
    \textbf{KO-OCR.} These are questions that simply ask the VLM to extract Korean text from an image. This evaluates the model's ability to read Korean text written in Hangul, which is an important skill for understanding scenes involving Korean text.
    
\item \textbf{Reasoning}
    
    \textbf{Commonsense Reasoning.} These questions ask for responses that require the model to reason about the scene presented in the image using commonsense knowledge and logic. This evaluates the model for its ability to understand a complex visual scene on a high level.

    \textbf{Document Understanding.} These are questions paired with images of documents with a large amount of Korean-language text, such as wiki articles, posters, and infographics. This evaluates the model for its ability to understand and reason about a large amount of Korean-language text presented visually.

    \textbf{Table Understanding.} These are questions paired with images of tables containing structured information in the Korean language. This evaluates the model for its ability to understand and reason about structured visual data.

    \textbf{Graph and Chart Understanding.} These are questions paired with images of various graphs and charts in the Korean language. This evaluates the model for its ability to understand and reason about graphs and charts that display information visually.

\item \textbf{Safety and Bias}

    \textbf{Hallucination and Robustness.} These are questions designed to potentially induce language-based hallucinations in a VLM by using unusual images. This evaluates the model for its ability to produce correct output based on visual information instead of hallucinating.
    
\end{enumerate}

The number of questions in each subcategory is provided in \Cref{tab:amount}. The above categories are constructed based on how we can categorize aspects of a VLM's performance that are unique to VLMs in that they relate to the model's ability to perceive and utilize visual input. Such an ability requires that it 1) can correctly perceive scenes and objects, and 2) is able to extract meaning from them. These each correspond to the Perception and Reasoning categories respectively. In addition to the above, the tendency to hallucinate is an important aspect of performance specific to VLMs~\cite{liu2024survey} that is useful to measure on its own. As this can involve aspects of both perception and reasoning but is distinct from either, we set aside a third category, Safety and Bias, containing questions to evaluate the tendency for VLMs to hallucinate.

We separate the Recognition-KO subcategory from Recognition due to the fact that the ability to recognize visual subjects unique to Korean culture is an important aspect of performance when considering real-world usage of the Korean language. Thus, we opt to evaluate it as a separate subcategory. Additionally, when evaluating the ability to extract written text from visual input in the KO-OCR subcategory, we only consider Korean-language text written in Hangul. We do not include questions that prompt the VLM to extract non-Korean text in our benchmark, as such questions would not meaningfully evaluate its performance related to the Korean language.
Similarly, we also only include images containing Korean-language text in the Document Understanding, Table Understanding, and Graph and Chart Understanding subcategories.

\subsection{Evaluation}

The evaluation process of KOFFVQA employs the LLM-as-a-judge approach, where the model is prompted to evaluate responses based on pre-defined grading criteria.
The VLM responses are generated by prompting the model with an image followed by the corresponding question. Since the benchmark allows free-form responses, the VLM's output is unrestricted, enabling the model to generate its own responses without predefined constraints. The response is then fed to the LLM judge along with the predefined grading criteria, ensuring that the LLM judge is guided by these criteria to generate a score between 0 and 10 in an easily parsable format.
Aside from the LLM judgment process, the language of the response is also determined using langid~\cite{lui-baldwin-2012-langid}, and the final score is set to zero if the response is not in the Korean language. An exception is when the response entirely consists of numbers and special characters, as this should be considered a valid response even if the language cannot be determined. The process of filtering by language in this manner is necessary because a model that responds in a different language despite not being instructed to when asked a question in Korean cannot be said to be performing the intended task.
After the final scores for each question are extracted, they are averaged both per subcategory and over the entire benchmark. The average scores are multiplied by 10 in order to shift the range of scores to be between 0 and 100 for more intuitive interpretation and easier comparison with other benchmarks.

\subsection{Data Construction}

\textbf{Human Annotation.} For each sample, a human annotator selects images from various online sources based on each subcategory. Images depicting general scenes are sourced from the Open Images v7 dataset~\cite{benenson2022colouring}. Images for the KO-OCR subcategory are obtained from the KAIST Scene Text dataset~\cite{jung2011touch}, and some images for the Document Understanding, Table Understanding, and Graph and Chart Understanding subcategories are sourced from the Open Data Portal of the Government of the Republic of Korea. Also, some of the images for the Hallucination and Robustness subcategories are taken from HallusionBench~\cite{guan2024hallusionbench}.

Based on the selected image and the given subcategory, the annotator creates a question and its corresponding grading criteria. Each criterion is paired with a partial score such that an entirely correct response would result in a score of 10. The criteria written for each question and the partial scores assigned to each criterion are determined based on the following:

\begin{enumerate}
\item If there is only one part to the intended answer, and the response can only be either correct or incorrect, there is only one criterion and it is assigned 10 points.
\item When there are multiple parts to the question, or if the response can have varying degrees of correctness, there are multiple criteria. In this case, partial scores are assigned to each criterion based on how important it is in determining overall correctness.
\begin{enumerate}
    \item When one part of the question is harder to answer than another, that part is assigned a larger score.
    \item An exception to the above is when the harder part of a question is a relatively minor detail, and a response can be considered mostly correct even if the harder part is incorrect. In this case, the harder part is assigned a smaller score.
    \item When a question can be answered with varying degrees of accuracy, there can be multiple mutually exclusive criteria for each possibility. In this case, the most accurate response is assigned a perfect score of 10, and less accurate responses are assigned a lower score.
\end{enumerate}
\end{enumerate}

Each of the grading criteria is written as an assessment of whether or not a particular answer or part of an explanation is included in a response. This way, responses generated by VLMs can be objectively judged based on their contents, minimizing the need for the judge model to subjectively interpret the criteria. This, in turn, allows for even small open-source models to be used as a judge for reliable evaluation, making the evaluation process more accessible compared to most LLM-as-a-judge approaches.

\noindent\textbf{Manual Curation.} In the process of creating questions and criteria for our benchmark, we manually edit and filter the data to create a meticulously curated benchmark well-suited for machine evaluation. After creating a set of questions, we use it to evaluate several VLMs and observe the generated judgments from our judge model. From this, we identify questions that the judge model has trouble correctly grading. This allows us to either edit the criteria so that it accounts for the possibility of the observed misjudgment, or remove the question entirely. This process is iteratively repeated to finally produce the 275 questions and grading criteria in our benchmark.

While KOFFVQA consists of human-written and well-curated questions and grading criteria in order to be as effective as possible, we believe our approach can also be easily scaled up by either using synthetic data for both questions and grading criteria or by generating grading criteria from existing questions using pre-trained models.
\section{Experiments and Results}
\label{sec:experiments}
\begin{table*}[t]
\centering
\begin{adjustbox}{width=\linewidth}
\begin{tabular}{rlcccccccccccc}
\hline
\multirow{2}{*}{Rank} & \multirow{2}{*}{Name} & Params & Overall & Obj. & \multirow{2}{*}{Recog.} & Recog. & \multirow{2}{*}{Relat.} & \multirow{2}{*}{KO-OCR} & Comm. & Doc. & Table & Graph \& Chart & Halluc. \& \\
 & & (B) & Score & Attr. & & -KO & & & Reas. & Underst. & Underst. & Underst. & Robust. \\
 \hline
 1 & \verb|gpt-4o-2024-11-20| & & 82.0 & 78.3 & 90.0 & 85.0 & 80.0 & 91.5 & 85.6 & 86.7 & 82.3 & 74.7 & 70.0 \\
 2 & \verb|claude-3-5-sonnet-20241022| & & 80.5 & 81.8 & 90.0 & 80.0 & 66.0 & 76.5 & 88.9 & 78.0 & 73.7 & 88.7 & 80.0 \\
 3 & \verb|Qwen2.5-VL-72B-Instruct| & 73.4 & 80.1 & 81.8 & 90.0 & 45.0 & 63.7 & 95.0 & 85.3 & 83.3 & 84.0 & 76.7 & 80.0 \\
 4 & \verb|gemini-2.0-flash-001| & & 79.1 & 73.5 & 80.0 & 70.0 & 65.3 & 93.5 & 72.7 & 90.0 & 96.7 & 88.0 & 50.0 \\
 5 & \verb|gemini-2.0-flash-exp| & & 78.9 & 73.8 & 80.0 & 70.0 & 56.7 & 90.0 & 82.7 & 93.3 & 90.0 & 84.7 & 50.0 \\
 6 & \verb|gpt-4o-2024-08-06| & & 77.6 & 77.5 & 80.0 & 90.0 & 64.7 & 80.0 & 87.6 & 77.0 & 82.0 & 68.0 & 70.0 \\
 7 & \verb|gemini-2.0-pro-exp-02-05| & & 77.6 & 80.0 & 80.0 & 90.0 & 58.0 & 85.0 & 87.1 & 86.0 & 78.7 & 68.0 & 50.0 \\
 8 & \verb|gemini-1.5-pro-002| & & 77.2 & 71.3 & 90.0 & 60.0 & 69.3 & 62.5 & 83.3 & 94.7 & 80.0 & 84.7 & 60.0 \\
 9 & \verb|Qwen2-VL-72B-Instruct| & 73.4 & 74.8 & 86.7 & 80.0 & 45.0 & 62.7 & 75.0 & 83.1 & 64.0 & 84.3 & 61.3 & 70.0 \\
 10 & \verb|gemini-1.5-flash-002| & & 73.5 & 72.8 & 90.0 & 50.0 & 68.0 & 72.5 & 78.2 & 89.3 & 83.3 & 61.3 & 40.0 \\
 11 & \verb|Ovis2-8B| & 8.9 & 69.5 & 77.8 & 65.0 & 30.0 & 70.7 & 70.0 & 76.2 & 53.7 & 74.0 & 64.0 & 80.0 \\
 12 & \verb|gpt-4o-mini-2024-07-18| & & 68.3 & 71.3 & 80.0 & 35.0 & 66.3 & 100.0 & 77.8 & 63.0 & 47.7 & 61.3 & 70.0 \\
 13 & \verb|Llama-3.2-90B-Vision-Instruct| & 88.6 & 67.9 & 75.0 & 75.0 & 40.0 & 62.7 & 80.0 & 76.0 & 56.0 & 55.0 & 68.0 & 76.0 \\
 14 & \verb|Qwen2.5-VL-7B-Instruct| & 8.3 & 67.7 & 78.3 & 50.0 & 25.0 & 55.3 & 85.0 & 71.6 & 74.7 & 61.7 & 58.7 & 75.0 \\
 15 & \verb|InternVL2_5-78B| & 78.4 & 67.2 & 71.3 & 75.0 & 25.0 & 66.7 & 70.0 & 78.9 & 60.7 & 51.0 & 68.0 & 85.0 \\
 16 & \verb|VARCO-VISION-14B| & 15.2 & 66.0 & 76.7 & 45.0 & 10.0 & 58.0 & 85.0 & 70.7 & 48.3 & 74.0 & 63.3 & 80.0 \\
 17 & \verb|gemini-2.0-flash-lite-preview-02-05| & & 65.2 & 59.8 & 75.0 & 60.0 & 43.7 & 68.0 & 72.7 & 80.3 & 71.7 & 63.3 & 60.0 \\
 18 & \verb|gpt-4-turbo-2024-04-09| & & 65.2 & 76.7 & 90.0 & 60.0 & 76.3 & 30.0 & 80.0 & 47.3 & 39.3 & 64.7 & 80.0 \\
 19 & \verb|Ovis2-34B| & 34.9 & 64.3 & 81.0 & 80.0 & 35.0 & 77.0 & 45.0 & 80.9 & 37.7 & 40.0 & 60.0 & 70.0 \\
 20 & \verb|Qwen2-VL-7B-Instruct| & 8.3 & 63.2 & 73.2 & 50.0 & 40.0 & 56.0 & 70.0 & 74.9 & 64.3 & 50.0 & 53.3 & 60.0 \\
 21 & \verb|gemini-1.5-flash-8b-001| & & 61.9 & 68.7 & 50.0 & 15.0 & 46.7 & 35.0 & 82.4 & 91.3 & 51.3 & 53.3 & 55.0 \\
\hline
\end{tabular}
\end{adjustbox}
\caption{Selected evaluation results for the top scoring 21 VLMs out of the 47 total models tested on our benchmark. A larger model size does not necessarily correspond to better performance, and models that excel in some subcategories may not do well in others. Due to the page limit, we show the entire model's result in \url{https://huggingface.co/spaces/maum-ai/KOFFVQA-Leaderboard}}
\vspace{-1em}
\label{tab:leaderboardil}
\end{table*}

\subsection{Evaluation of Existing Models}

In this section, we present the evaluation results of existing open- and closed-source VLMs using our benchmark and analyze the results. We evaluate a total of 47 VLMs. The models we evaluate are the following:

\noindent\textbf{VLMs that officially support Korean or explicitly mention using Korean data in training}: Claude models(3.5 Sonnet 20241022, 3 Haiku 20240307, 3 Sonnet 20240229, 3 Opus 20240229)~\cite{claude3,Claude3.5}, Gemini models(2.0 Pro experimental 02-05, 2.0 Flash 001, 2.0 Flash experimental, 2.0 Flash Lite preview 02-05, 1.5 Pro 002, 1.5 Flash 002, 1.5 Flash 8B 001)~\cite{team2024gemini}, GPT-4o models(2024-08-06, 2024-11-20, Mini 2024-07-18)~\cite{hurst2024gpt}, GPT-4V~\cite{achiam2023gpt}, Bllossom-AICA-5B~\cite{shin-etal-2024-x}, InternVL 2.5(8B, 26B, 38B, 78B)~\cite{chen2024expanding}, MiniCPM-V 2.6~\cite{yao2024minicpm}, Qwen2-VL(3B, 7B, 72B)~\cite{wang2024qwen2}, and VARCO-VISION 14B~\cite{ju2024varco}.

\noindent\textbf{VLMs that do not officially support Korean and do not mention using Korean data in training}: Aria~\cite{li2024aria}, Idefics 3 8B~\cite{laurenccon2024building}, InternLM-XComposer 2.5 7B~\cite{zhang2024internlm}, InternVL 2(8B and 76B)~\cite{chen2024far}, Llama 3.2(11B and 90B)~\cite{grattafiori2024llama}, LLaVA-OneVision(7B and 72B)~\cite{li2024llava}, MAmmoTH-VL 8B~\cite{guo2024mammoth}, Molmo(7B, 72B)~\cite{deitke2024molmo}, Ovis 2(8B and 34B)~\cite{lu2024ovis}, Ovis 1.6(9B and 27B)~\cite{lu2024ovis}, Phi 3.5 Vision~\cite{abdin2024phi}, Pixtral 12B~\cite{agrawal2024pixtral}, Qwen2.5-VL(2B, 7B, 72B)~\cite{bai2025qwen2}, and SmolVLM~\cite{marafioti2025smolvlm}.

For the judge LLM, we use Gemma 2 9B~\cite{team2024gemma} with the temperature set to zero for this experiment. More advanced proprietary models, such as GPT-4o and Gemini 2.0 Flash, can also be used. We mainly use Gemma 2 9B as it is an open-source model that 1) can be reasonably deployed on a local machine, and 2) can meaningfully understand the grading criteria given in Korean in a manner that aligns with human intention. Being an open-source model also makes the results easily reproducible. Gemma 2 9B correctly judges responses in approximately 89\% of all cases.

A subset of the evaluation results, ordered by overall score, is provided in \Cref{tab:leaderboardil}. At first glance, a notable observation we can make is that a larger model does not necessarily correspond to better performance, contrary to the tendency in English-language benchmarks. We suggest that this is because different models are trained on varying amounts of multilingual data, which can have an impact on performance, especially for less-represented languages such as Korean. Inspecting the responses of each model, we see that many models that excel in English often fail to answer simple questions in KOFFVQA by either generating incoherent Korean or answering the question in English even when prompted in Korean.

We can also observe that a high overall score does not necessarily imply a high score in every subcategory. In fact, we can observe that many models excel in some categories but not in others. This suggests that the categorization system of our benchmark is meaningful and can provide useful information about various aspects of a VLM's performance. For instance, the top five models in the Document Understanding subcategory are Gemini models, with scores ranging from 94.67 to 89.33. Even the smallest model, \verb|gemini-1.5-flash-8b-001|, which is ranked 21st overall, outperforms the overall top model(\verb|gpt-4o-2024-11-20|) in this subcategory.

\subsection{Comparison with Other Evaluation Methods}

In this section, we perform an experiment to verify that our method of judge model-based evaluation using pre-determined grading criteria is more robust and resistant to subjectivity and unreliability compared to existing methods of judge-based evaluation used in many VLM benchmarks.

\begin{table}[t]
\centering
\resizebox{\linewidth}{!}{
\begin{tabular}{cccc}
\hline
& Gemma 2 9B & GPT-4o & Gemini 2.0 Flash \\ \hline
KOFFVQA   & 0.398      & 0.171  & 0.127  \\
KOFFVQA-V &      -      & 0.208  & 0.254  \\
KOFFVQA-GT & 0.584      & 0.456  & 0.476  \\
KOFFVQA-GT-V &    -        & 0.452  & 0.426  \\ \hline
\end{tabular}
}
\caption{Mean standard deviation of scores for 5 repeated evaluations for each question across 275 sampled responses. These are based on individual scores that range from 0 to 10.}
\vspace{-1em}
\label{tab:sdcomp}
\end{table}

We create three alternate versions of our benchmark: 1) KOFFVQA-V, where the image belonging to each question is given to a VLM judge in addition to our grading criteria, 2) KOFFVQA-GT, where no image is given and a "ground truth" response is given instead of grading criteria, and 3) KOFFVQA-GT-V, where the "ground truth" response is given along with the image paired with the question. The "ground truth" responses we use are minimal natural language responses written by a human to satisfy all of the grading criteria for each question. We compare the evaluation results of these with the results from our original KOFFVQA benchmark.

We also compare the results of evaluation between three different judge models. We use Gemma 2 9B as a judge for KOFFVQA and KOFFVQA-GT, and also GPT-4o(2024-11-20), and Gemini 2.0 Flash as a judge for all four benchmarks. We select GPT-4o and Gemini 2.0 Flash to represent two different robust proprietary VLMs that can be employed as a judge for our benchmark. For this experiment, we set the generation temperature to 0.6 for Gemma 2 9B and 1.0 for GPT-4o and Gemini 2.0 Flash, and run the evaluation process 5 times over a fixed subset of 275 responses randomly sampled from the evaluation results of our tested VLMs. When creating this subset, we randomly choose the response of one model for each of the 275 questions. As we are only considering the judgments of the judge models and not the language filtering process, we exclude responses that are in a non-Korean language when we sample the responses. Our intention is to compare the reliability of the judge model for each grading method by measuring the standard deviations of scores across multiple trials.

A comparison of the standard deviations of scores for 5 trials averaged across all 275 samples is given in \Cref{tab:sdcomp}. In this experiment, we compare the differences in standard deviation between KOFFVQA and KOFFVQA-GT, and between KOFFVQA-V and KOFFVQA-GT-V.
We can first observe that for all 3 judge models, providing grading criteria(KOFFVQA, KOFFVQA-V) results in much more consistent evaluation results compared to providing a ground-truth answer(KOFFVQA-GT, KOFFVQA-GT-V). This is expected, as grading criteria provide an objective metric with which to grade the responses, as opposed to ground truth answers which do not specify how to grade a partially correct answer.

\begin{figure*}[t]
\includegraphics[width=0.99\linewidth]{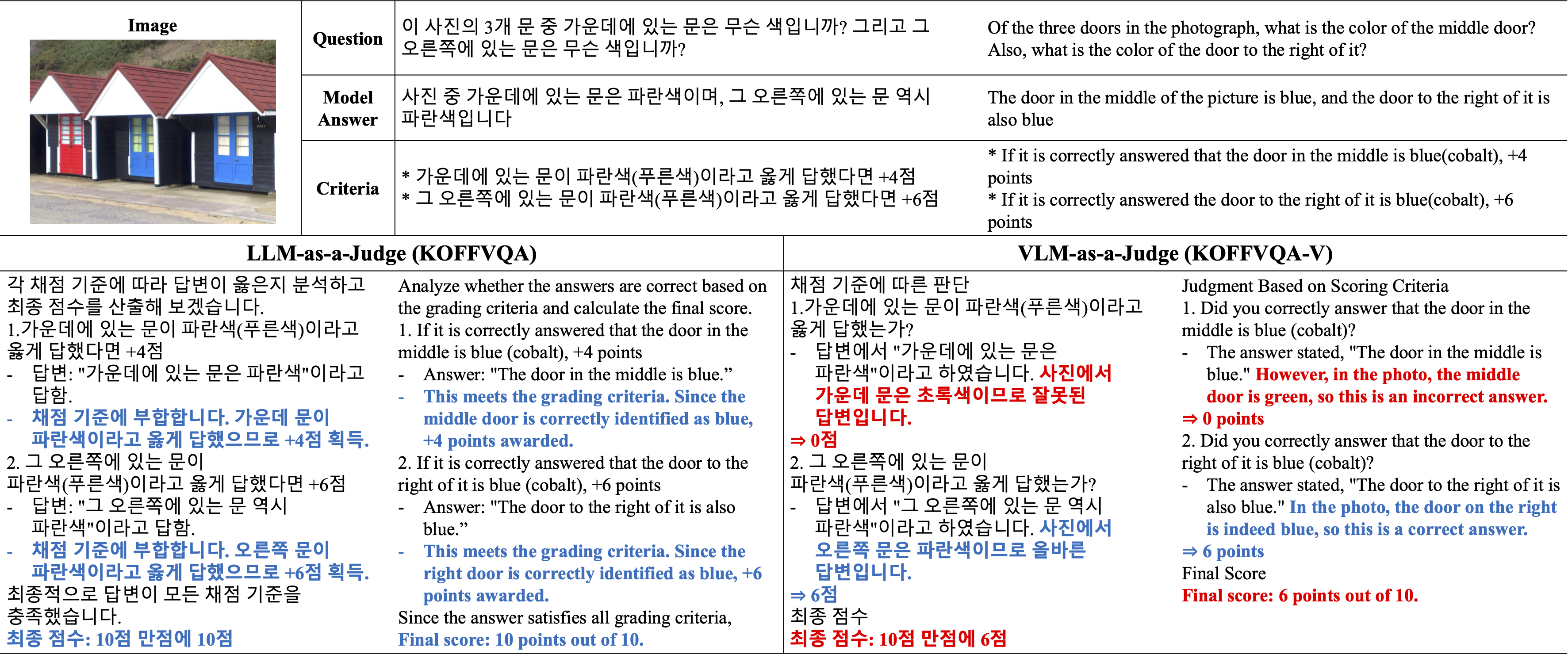}
\caption{An example of a response that GPT-4o grades \textcolor{blue}{correctly} when the image is not given as input but grades \textcolor{red}{incorrectly} when the image is given. The left columns are the original text in Korean, and the right columns provide the English translations. When the image is given, the judge model attempts to judge the response based on the image and hallucinates that the door in the middle of the photograph is green. When the image is not given, the judge has no reason to grade the response based on anything other than the given criteria.}
\vspace{-1em}
\label{tab:hallucinationexample}
\end{figure*}

\subsection{Influence of Visual Input in Judgment}

We also intend to prove that prompting the judge with image inputs can be less reliable compared to using only language inputs in judgment using our method due to hallucinations based on visual input.

An important observation that we can make is that when comparing KOFFVQA and KOFFVQA-V in \Cref{tab:sdcomp}, providing images in addition to grading criteria seems to degrade the consistency of evaluation for both GPT-4o and Gemini 2.0 Flash. This is despite the fact that the judges are given identical grading criteria for both benchmarks, and are prompted to grade the responses using only the grading criteria regardless of the contents of the image.

\begin{table}[t]
\centering
\resizebox{\linewidth}{!}{
\begin{tabular}{cccc}
\hline
& Gemma 2 9B & GPT-4o & Gemini 2.0 Flash \\ \hline
KOFFVQA   & 89.3\%     & 95.8\% & 93.0\% \\
KOFFVQA-V &  -         & 92.9\% & 91.8\% \\ \hline
\end{tabular}
}
\caption{Percentages of how often each judge model correctly graded the sampled responses according to the grading criteria. These values are derived by comparing the grading results to human-evaluated scores.}
\label{tab:correctness}
\vspace{-1em}
\end{table}

We also compare the accuracy of how often each judge model correctly graded the responses using the given criteria when compared to human-evaluated scores. As the criteria provide objective guidelines for grading, true scores for each of the 275 sampled responses can be acquired by manually grading each of the responses using the existing grading criteria. The correctness of evaluation for each judge model in KOFFVQA and KOFFVQA-V are given in \Cref{tab:correctness}.
The grading correctness decreases when an additional image input is given, despite the judge models in both cases having been given identical grading criteria.

We further explore the cause of this phenomenon by qualitatively comparing the judgments of each model when grading the same response in both KOFFVQA and KOFFVQA-V. An example of a case where GPT-4o grades a response correctly in the KOFFVQA case but incorrectly in the KOFFVQA-V case is shown in \Cref{tab:hallucinationexample}. In this example, when grading the response with only the grading criteria, the judge correctly determines that the answer satisfies both criteria and gets 10 points. On the other hand, when the judge is also given the image input, it attempts to base its judgment on the image and the grading criteria. Since it hallucinates that the middle door is green instead of blue, it incorrectly determines that the response does not satisfy the first grading criterion. Many other samples where the same judge graded the KOFFVQA case correctly but misgraded the KOFFVQA-V case exhibited a similar pattern, where the judge hallucinated details of the image and evaluated the response based on this information instead of the given criteria. This shows that the tendency of VLMs to hallucinate based on visual input negatively impacts the grading performance of judge models when used in a benchmark that utilizes our approach of using grading criteria. This confirms that there is an issue of VLM judges producing incorrect judgments due to hallucination prompted by visual input. The KOFFVQA benchmark does not suffer from this issue as it uses the LLM-as-a-judge method.

\subsection{Analysis of failure cases}

During the grading process of the 275 sampled questions in the above experiment, we inspect cases where the judge models graded a response differently from the human-evaluated score in the KOFFVQA benchmark.
We manually review 11 failure cases of GPT-4o and 17 failure cases of Gemini 2.0 Flash from 275 grading results of the original subset. We do not analyze the judgments from Gemma 2 9B, as the model does not output the reasoning behind its evaluation the majority of the time. Based on the judge model's reasoning, we classify the failure cases into the following 3 categories:

\begin{enumerate}
\item \textbf{Arbitrarily overruling criteria}. This is when the judge model decides to overrule what is explicitly written in the grading criteria and judge the response based on its own interpretation of the criteria.
\item \textbf{Ambiguous criteria}. This is when the grading criteria are not perfectly clear in meaning in the context of the provided response, and the judge model misinterprets the grading criteria due to this ambiguity.
\item \textbf{Unexplainable misjudgment}. This is when the judge incorrectly grades the response for no apparent reason.
\end{enumerate}

\noindent GPT-4o has 5 cases in category 1, 2 cases in category 2, and 4 cases in category 3. Gemini 2.0 Flash has 7 cases in category 1, 2 cases in category 2, and 8 cases in category 3. We see that the majority of failure cases is in categories 1 and 3, suggesting that there is even more room for improvement in our LLM-as-a-judge process if a judge model can be trained to minimize these errors.
\section{Conclusion}
\label{sec:conclusion}

In this work, we identify an issue with existing benchmarks for the evaluation of large vision-language models, namely that benchmarks leveraging the open-ended generation ability of VLMs suffer from the issue of unreliable judgment in the evaluation process. In addition, we observe that there is a lack of general-purpose benchmarks for VLMs in the Korean language that also include Korea-specific content. To address these issues, we develop KOFFVQA, a carefully crafted general-purpose free-form Korean language VQA benchmark using a novel approach of providing objective pre-defined grading criteria to an LLM judge model during the evaluation process.

We use our benchmark to evaluate the performance of 47 VLMs, including both open and closed models, regardless of their support for Korean, and find it effective in evaluating various aspects of each model's performance. We also experimentally compare our partial scoring approach with the existing method of prompting the judge to compare each response with a reference response and show that our approach results in a more reliable and consistent judgment. We also demonstrate that using a VLM judge with image input can degrade grading effectiveness and cause hallucinations. Furthermore, we analyze the failure cases of LLM judge models using our benchmark and suggest that the accuracy of our judge model can be improved by training.

We hope that KOFFVQA can provide insights on the performance of various VLMs in the Korean language and serve as a reference for the further development of other benchmarks that utilize fine-grained criteria in the future.

{
    \small
    \bibliographystyle{ieeenat_fullname}
    \bibliography{main}
}


\end{document}